\documentclass[conference]{IEEEtran}
\usepackage{cite}
\usepackage{amsmath,amssymb,amsfonts}
\usepackage{algorithm}
\usepackage{algorithmic}
\usepackage{graphicx}
\usepackage{textcomp}
\usepackage{xcolor}
\usepackage[hyphens]{url}
\usepackage{fancyhdr}
\usepackage{hyperref}
\usepackage{array}
\usepackage{tabularx}
\usepackage{bm}
\usepackage{booktabs}
\usepackage{authblk}
\usepackage[switch]{lineno}
\usepackage{multirow}
\usepackage{adjustbox}
\usepackage{float}
\usepackage{makecell, hhline}

\def\BibTeX{{\rm B\kern-.05em{\sc i\kern-.025em b}\kern-.08em
    T\kern-.1667em\lower.7ex\hbox{E}\kern-.125emX}}

\def\invcircledast#1{%
  \mathbin{\vphantom{\circledast}\text{%
    \ooalign{\smash{\blackcircle}\cr
             \hidewidth\smash{\textcolor{white}{\bf \footnotesize $#1$}}\hidewidth\cr
            }%
  }}%
}
\newcommand{\blackcircle}{\raisebox{-.6ex}{\scalebox{2.30}{$\bullet$}}}

\newcommand{\NeuroHash}{\emph{NeuroHash} }

\title{NeuroHash: A Hyperdimensional Neuro-Symbolic Framework for Spatially-Aware Image \\Hashing and Retrieval} 
\author[1]{Sanggeon Yun}
\author[1]{Ryozo Masukawa}
\author[1]{SungHeon Jeong}
\author[1, $\dag$]{Mohsen Imani}
\affil[1]{University of California, Irvine}
\affil[$\dag$]{Corresponding author, email: m.imani@uci.edu}

\begin{document}
\maketitle
\thispagestyle{plain}
\pagestyle{plain}

\begin{abstract}
    Customizable image retrieval from large datasets remains a critical challenge, particularly when preserving spatial relationships within images. Traditional hashing methods, primarily based on deep learning, often fail to capture spatial information adequately and lack transparency. In this paper, we introduce \NeuroHash, a novel neuro-symbolic framework leveraging Hyperdimensional Computing (HDC) to enable highly customizable, spatially-aware image retrieval. \NeuroHash combines pre-trained deep neural network models with HDC-based symbolic models, allowing for flexible manipulation of hash values to support conditional image retrieval. Our method includes a self-supervised context-aware HDC encoder and novel loss terms for optimizing lower-dimensional bipolar hashing using multilinear hyperplanes. We evaluate \NeuroHash on two benchmark datasets, demonstrating superior performance compared to state-of-the-art hashing methods, as measured by mAP@5K scores and our newly introduced metric, mAP@5K$_r$, which assesses spatial alignment. The results highlight \NeuroHash's ability to achieve competitive performance while offering significant advantages in flexibility and customization, paving the way for more advanced and versatile image retrieval systems.

\end{abstract}

\section{Introduction}

In the era of explosive growth in image data, managing vast repositories of images, particularly in domains requiring the swift retrieval of similar images for a given query, presents an escalating challenge. Numerous research endeavors have sought to develop efficient and accurate methods for similar image retrieval. The primary focus of these investigations has been the design of adept hash functions capable of transforming images into a compact, fixed-size hash, thereby encapsulating their similarity to other images.

One early research utilized shallow machine learning models such as support vector machine (SVM) to extract discrete features from each image to hashing images~\cite{shen2015supervised}. As deep neural networks (DNNs) show remarkable performance on various image-based tasks, image hashing models for image retrieval based on neural networks are proposed starting from Convolutional Neural Networks (CNNs) based approaches~\cite{xia2014supervised,cao2017hashnet}. 
This momentum of applying DNNs to image retrieval tasks evolved towards purely attention mechanism-based models~\cite{chen2021crossvit}.
Nowadays, state-of-the-art models on image retrieval tasks only exhibit end-to-end DNN-based architectures.

Despite the strides made by previous deep hashing-based methods using gradient-based end-to-end deep learning models with specifically designed loss functions to capture global and local information of images, this black-box training approach does not guarantee the embedding of desired information, including local or spatial details. Furthermore, because these end-to-end deep learning models are trained with predetermined criteria, they have fundamental limitations in conducting image retrieval flexibly, such as with additional conditions like precise positioning of each object or prioritizing specific objects during image retrieval.

\begin{figure}
    \centering
    \includegraphics[width=1.\linewidth]{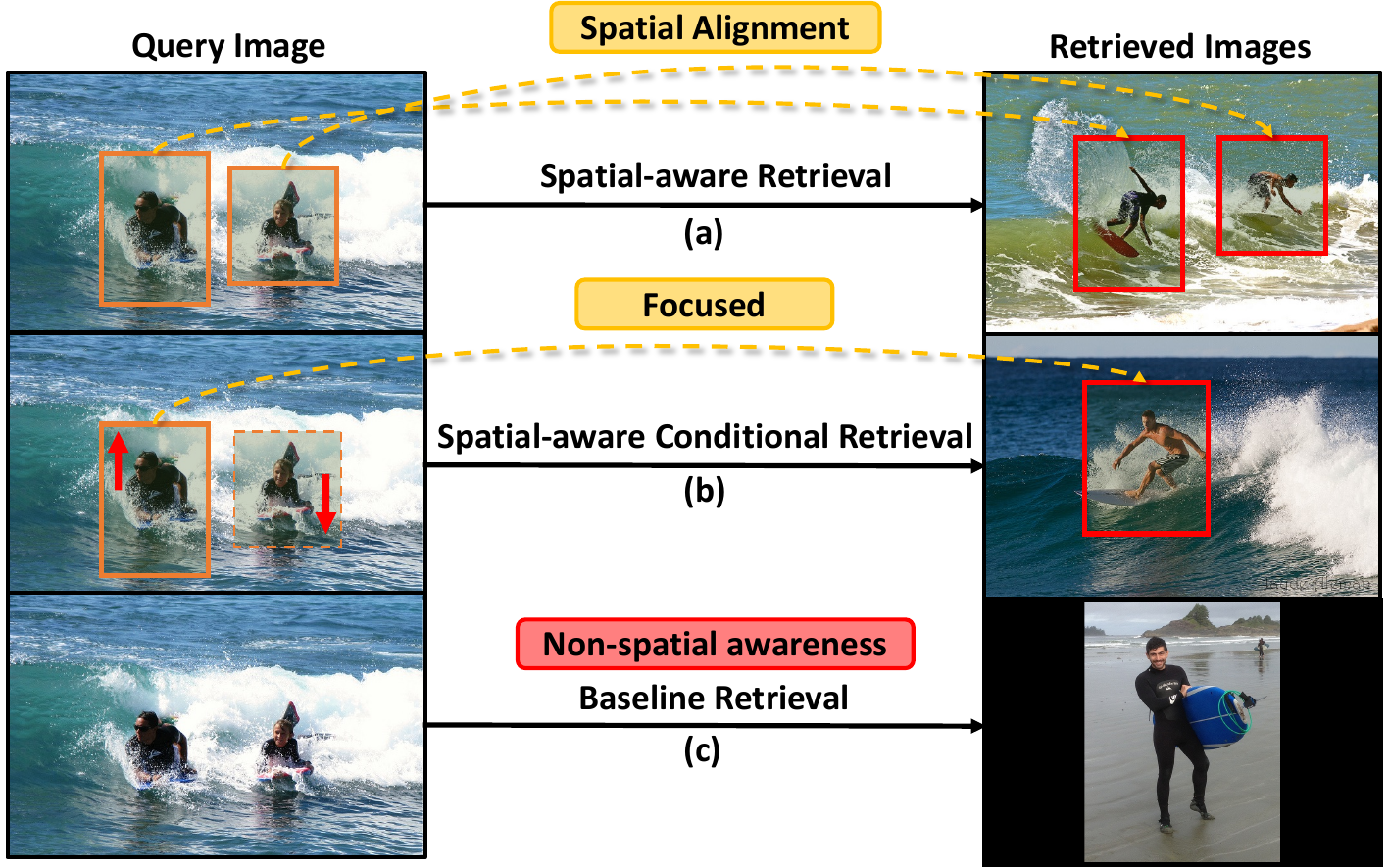}
    \vspace{-5mm}
    \caption{\textbf{The actual image retrieval results comparing our framework's (a) spatial-aware retrieval and (b) conditional retrieval with (c) baseline retrieval}.}\label{fig:NeuroHash_overview}
    \vspace{-6mm}
\end{figure}

To resolve the above limitations of previous methods, we propose an innovative image hashing method employing Hyperdimensional Computing (HDC)~\cite{kanerva2009hyperdimensional} to facilitate image retrieval with spatial structural conditions that can be easily manipulated as illustrated in~\autoref{fig:NeuroHash_overview}. HDC stands as an alternative paradigm inspired by essential brain functions, emphasizing high efficiency and symbolic learning capabilities. Recently, HDC has demonstrated the power of combining with neural network models~\cite{hersche2023neuro}. Building on the success of neuro-symbolic AI with HDC and the observation that the human brain excels at manipulating high-dimensional representations, our approach harnesses HDC operations to embed spatial structural information into a high-dimensional vector in a neuro-symbolic manner, constituting a hashed representation of the image.

Our methodology capitalizes on pre-trained large vision models to extract feature representations for individual objects, subsequently combining them into a singular representation of high-dimensional vectors through spatial encoding – a process applying HDC operations with positional information. These representations are then hashed into lower-dimensional bipolar vectors to facilitate rapid image retrieval. During the retrieval process, our method replicates the spatial encoding procedure to retrieve images with similar spatial structures. Additionally, structural conditions can be controlled by incorporating HDC operations on a given query image, such as focusing on spatial information of a specific object.

In summary, our work represents a fundamentally novel contribution to the field, offering the following key advancements:

\begin{itemize}
    \item We propose \NeuroHash, a novel self-supervised neuro-symbolic image hashing framework designed to enable customizable spatial-aware image retrieval. Unlike previous methods that rely on fully gradient-based end-to-end deep learning models, our approach combines DNN-based models with HDC-based symbolic models to symbolically encode spatial information with local features, allowing for flexible manipulation of hash values and enhancing interpretability.  This enables conditional image retrieval that can focus on specific objects or spatial regions within an image, offering a high degree of customization and control.
    \item We introduce a hyperdimensional spatial encoding technique, which is, to the best of our knowledge, the first HDC encoding method that preserves spatial similarity.
    \item We devise a context-aware HDC encoder that preserves characteristics from the original feature space using self-supervised training, advancing previous HDC encoders.
    \item We propose new loss terms for lower-dimensional bipolar hashing using multilinear hyperplanes to enhance hash function optimization.
    \item Experimental results on two benchmark datasets show that \NeuroHash outperforms state-of-the-art hashing methods in terms of mAP@5K scores. These results demonstrate that our neuro-symbolic framework can achieve performance comparable to fully gradient-based end-to-end models while offering additional benefits in terms of flexibility and customization.
    \item We introduce a new metric, mAP@5K$_r$, to evaluate the effectiveness of spatial-aware and conditional image retrieval, demonstrating the framework's capability to align retrieved images with spatial constraints accurately.
\end{itemize}

\section{Related Works}

\subsection{Hash-based Approximate Nearest Search}

Retrieving similar vectors efficiently from abundant vector data using linear search or traditional structures is impractical. To address this, studies explore converting high-dimensional vectors into fixed-size, low-dimensional representations, with Locality Sensitive Hashing (LSH)\cite{gionis1999similarity} being a notable unsupervised algorithm\cite{shen2018unsupervised}. LSH constructs a hash table using multiple functions capturing local similarity, and variations like Multilinear Hyperplane Hashing~\cite{liu2016multilinear} specifically preserve cosine similarity in the hash space.

\subsection{Deep Hashing Approach}

In early image retrieval, methods such as supervised discrete hashing (SDH)\cite{shen2015supervised} played a crucial role in reducing storage and improving retrieval speed. The integration of Convolutional Neural Networks (CNNs) brought advancements with models such as 
HashNet~\cite{cao2017hashnet}, building on architectures like AlexNet~\cite{krizhevsky2012imagenet} 
to address discrete optimization challenges.

The evolution shifted towards deep learning, leveraging ResNet as a popular backbone network in approaches like CSQ~\cite{yuan2020central} and DBDH~\cite{zheng2020deep}. As models progressed, attention turned to hybrid models like DAgH~\cite{chen2019deep} and DAHP~\cite{li2021dahp}, using attention networks to enhance performance without increasing convolution layers. Scalability concerns led to exploration of a self-attention-based structure~\cite{chen2021crossvit}.

Recent developments expanded unsupervised deep hashing into applications like image copy detection~\cite{liu2019efficient} and image quality assessment~\cite{huang2020perceptual}. Methods like DeepBit~\cite{lin2018unsupervised}, DistillHash~\cite{yang2019distillhash}, and TBH~\cite{shen2020auto} explored unsupervised learning with novel loss functions. Contrastive learning in computer vision paved the way for unsupervised hashing methods such as HAMAN~\cite{ma2022improved} and MeCoQ~\cite{wang2022contrastive},
leveraging contrasting positive and negative samples for robust hash codes.


Certain unsupervised hashing methods focused on mining pairwise similarity. DistillHash~\cite{yang2019distillhash} and SSDH~\cite{yang2018semantic} used data pair distillation and semantic structures, while FSCH~\cite{cao2023unsupervised} extended these approaches with fine-grained similarity structures based on global and local image representations. Our proposed approach takes a completely different direction by leveraging an HDC-enhanced neuro-symbolic approach while following the previous strategy that combines local and global features.

\subsection{Hyperdimensional Computing}

Brain-inspired hyperdimensional computing (HDC) is based on the understanding that brains compute with patterns of neural activity that are not readily associated with numbers. Due to the huge size of the brain’s circuits, neural patterns can be modeled with hypervectors~\cite{kanerva2009hyperdimensional}. HDC builds upon a well-defined set of operations with random hypervectors, is extremely robust in the presence of failures, and offers a complete computational paradigm that is easily applied to multiple learning problems, such as speech recognition~\cite{imani2017voicehd}, graph learning~\cite{kang2022relhd,nunes2022graphhd}, and computer vision~\cite{hersche2022constrained,dutta2022hdnn}.

Recent literature has witnessed a growing interest in hyperdimensional computing (HDC) as a learning model, praised for its simplicity and computational efficiency. However, conventional HDC frameworks encounter issues with randomly generated and static encoders, leading to an abundance of parameters and decreased accuracy. LeHD~\cite{duan2022lehdc}, an innovative approach, employs a principled learning approach to refine model accuracy, transforming the HDC framework into an equivalent binary neural network architecture. These advancements collectively aim to overcome issues with static encoders in HDC, offering a more effective and accurate learning framework.

\section{Methodology}

\subsection{HDC Basics}\label{sec:symbolictraining}
The core of HDC is called a hyperdimensional vector, denoted $\mathcal{H}$, which represents a vector in $\mathbb{R}^D$ with a high dimensionality of $D$. Hyperdimensional vectors are compared using a similarity function $\delta$. By using this similarity measure, HDC becomes a versatile tool for cognitive tasks, including memory, classification, clustering, etc. HDC frameworks designed to support these tasks are based on three core operations that mirror brain functionalities: bundling, binding, and permutation. Here are the details of each operation:

\begin{enumerate}
    \item \textbf{Bundling}: This operation, represented by $+$, is commonly executed as element-wise addition. If $\mathcal{H}=\mathcal{H}_1+\mathcal{H}_2$, then both $\mathcal{H}_1$ and $\mathcal{H}_2$ exhibit similarity to $\mathcal{H}$. In terms of cognitive interpretation, this operation can be understood as a form of memorization.
    
    \item \textbf{Binding}: This operation, denoted by $*$, is usually implemented as an element-wise multiplication. If $\mathcal{H}=\mathcal{H}_1 * \mathcal{H}_2$, then $\mathcal{H}$ is dissimilar to both $\mathcal{H}_1$ and $\mathcal{H}_2$. Binding has a crucial property of similarity preservation, where for some hypervector $\mathcal{V}$, $\delta(\mathcal{V}*\mathcal{H}_1,\mathcal{V}*\mathcal{H}_2)\simeq\delta(\mathcal{H}_1,\mathcal{H}_2)$. From a cognitive point of view, this operation can be understood as an association. Binding can be used to associate different pieces of information, such as coordinates and image feature vectors, in hyperdimensional space.
    
    \item \textbf{Permutation}: This operator, represented by $\rho$, is commonly executed as a rotation of vector elements. In general, $\delta(\rho(\mathcal{H}),\mathcal{H})\simeq 0$. Permutation is frequently employed to encode the order within sequences.
\end{enumerate}

Leveraging the three fundamental HDC operations provides a foundation for a hyperdimensional learning framework applicable to various tasks. In the context of classification, each step of the framework can be outlined as follows.

\begin{enumerate}
    \item \textbf{Encoding}: The initial step within the HDC framework involves mapping the input data $\vec{F}\in U$ into a high-dimensional space through the introduction of an encoding function $\vec{\phi}:U\to H$, commonly known as \emph{encoding}. Consider an input vector with $n$ features, denoted as $\vec{F} = \{f_i\}$, representing features extracted from an image. The commonly used encoding function is defined as $\vec{\phi}(\vec{F})=\cos{(\vec{F}\times\vec{B}+\vec{b})}\times\sin{(\vec{F}\times\vec{B})}$, where $\vec{B}$ is an $n\times D$ matrix, and each element in $\vec{B}$ is sampled from an i.i.d Gaussian distribution with parameters ($\mu=0, \sigma=1$). Additionally, $\vec{b}$ is sampled from an i.i.d uniform distribution over the interval $[0, 2\pi]$. The $\vec{\phi}$ function preserves a notion of similarity in the input space. Consequently, for any given inputs $\vec{x}_1, \vec{x}_2\in U$, their corresponding hypervectors, $\vec{\phi}(\vec{x}_1)$ and $\vec{\phi}(\vec{x}_2)$, exhibit similarity iif $\vec{x}_1$ is similar to $\vec{x}_2$. Such initialized encoders with parameters $\vec{B}$ and $\vec{b}$ can be further optimized by making $\vec{B}$ and $\vec{b}$ learnable parameters using a gradient descent approach.
    
    \item \textbf{Symbolic Training}: Consider a dataset $\mathcal{D}\subset U$ where each data point $\vec{x}_i\in D$ is associated with a label $1 \leq y_i \leq m$ from a set of $m$ classes. In traditional hyperdimensional classifier training, the process involves generating $m$ class hypervectors through bundling: $\vec{C}_i = \sum_{y_j=i}{\vec{\phi}(\vec{x}_j)}$. For each data point to retrain $\vec{x}_i$, each class hypervector is updated as follows:
    
    $$
    \begin{aligned}
      \vec{C}_l & \leftarrow \vec{C}_l + \eta(1-\delta)\vec{\phi}(\vec{x}_i)\\
      \vec{C}_{l'} & \leftarrow \vec{C}_{l'} - \eta(1-\delta)\vec{\phi}(\vec{x}_i)
    \end{aligned}
    $$
    where $l=y_i, l'\neq y_i$, $\delta=\delta(\vec{C}_l, \vec{\phi}(\vec{x}_i))$, and $\eta$ is learning rate.

    \item \textbf{Symbolic Inference}: Once the class hypervectors $\vec{C}_i$ undergo updates through the initial training phases, the classification of a given query $\vec{q}\in D$ becomes a straightforward process. A class $i$ is predicted when $\delta(\vec{C}_i, \vec{\phi}(\vec{q})) > \delta(\vec{C}_j, \vec{\phi}(\vec{q}))$ is satisfied for all $j\neq i$.
\end{enumerate}

\subsection{Proposed Framework}

\begin{figure*}[!t]
    \centering
    \includegraphics[width=1\textwidth]{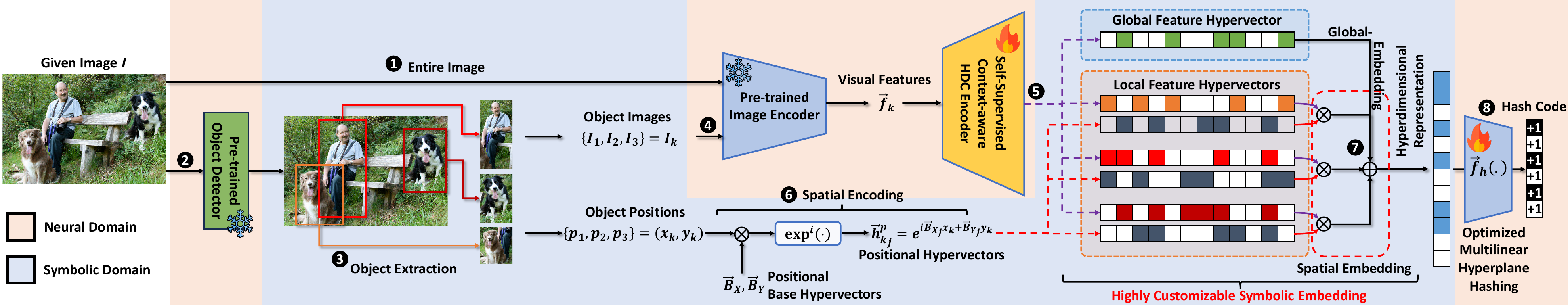}
    \vspace{-7mm}
    \caption{\textbf{Overall pipeline of our proposed \NeuroHash a novel neuro-symbolic framework for spatial-aware hashing and conditional image retrieval in a self-supervised manner}.}\label{fig:pipeline}
    \vspace{-6mm}
\end{figure*}

\subsubsection{Overall Pipeline}

The overall pipeline of our proposed framework is presented in~\autoref{fig:pipeline}. First, given an image $I$, we extract global features, which is embedding of the image, through a pre-trained image encoder model by giving the entire given image to the model ($\invcircledast{1}$). Also, in order to consider local information, it extracts bounding boxes indicating objects that are presented in the image by conducting an object detection task over the image using a pre-trained object detection model ($\invcircledast{2}$). Using the bounding boxes that are generated during the previous step, it extracts two types of object information: object images $I_k$ and object positions $p_k=(x_k, y_k)$ ($\invcircledast{3}$). With the extracted object images $I_k$, visual features $\vec{f}_k$ corresponding to each of the object images $I_k$ are computed by using the same pre-trained image encoder model that is used during the global features extraction ($\invcircledast{4}$). The resulting visual features including global and local information are sent to an HDC encoder to have visual feature hypervectors ($\invcircledast{5}$). Not only considers the local visual information, to further consider local spatial information, but it also conducts spatial encoding using two positional base hypervectors each corresponding to $x$ and $y$ coordinate, and computes positional hypervectors ${\vec{h}^{p_k}}_j\in \mathbb{C}$ ($\invcircledast{6}$). Finally, it combines local visual features with local positional features to have spatial embedding addition to that, it also combines global feature hypervector to have global embedding ($\invcircledast{7}$). After the final hypervector embedding process, hyperdimensional representation corresponding to the given image $I$ is generated. The generated hyperdimensional representation is now hashed into a compact binary hash representation with the optimized multilinear hyperplane hashing model $\vec{f}_h(.)$ ($\invcircledast{8}$).

\subsubsection{Global and Local Visual Features Extraction}

In an image retrieval task, it is crucial to well-represent each image in a compact representation. Although pre-trained large image embedding models introduced so far present powerful performance in extracting visual features, simply embedding entire images can lead to insufficient interpretation of local information considering the complexity of image data. To allow solid local visual information consideration, we propose to employ a pre-trained object detection model in order to extract objects that are presented in a given image. Therefore, our proposed framework uses two pre-trained large image models: 1) object detection model $f_{obj}\colon \mathcal{I}\to \mathcal{B}, \mathcal{B}\subseteq \mathbb{R}^{N\times 4}$ where $\mathcal{I}$ indicates input image space and $\mathcal{B}$ indicates a set of bounding boxes each contains position and the size of the box and, 2) image embedding model $\vec{\phi}_{vis}\colon \mathcal{I} \to \mathcal{Z}, \mathcal{Z}\subseteq \mathbb{R}^z$ where $\mathcal{Z}$ indicates embedding space with $z$ dimensionality. First, global visual feature vector $\vec{f}_{glob}\in\mathbb{R}^z$ is easily retrieved by passing entire image $I$ to $\vec{\phi}_{vis}$. In the case of the local visual features extraction, $f_{obj}$ needs to be utilized before retrieving features. By applying object detection $f_{obj}$ over the given image $I\in\mathcal{I}$, we can obtain $N$ bounding boxes $BB_k\in\mathbb{R}^{4}$ that correspond to each object in the image. Using each $BB_k$, local visual features $\vec{f}_k$ can be retrieved by applying cropped images $I_k$ based on $BB_k$ to $\vec{\phi}_{vis}$.

\subsubsection{Context-aware HDC Encoding}

Inspired by the previous work LeHD~\cite{duan2022lehdc}, we designed an HDC encoder $\vec{\phi} \colon \mathbb{R}^z \to \mathbb{R}^D$ ($D\gg z$) that is capable of encoding visual features into a feature hypervector $\vec{h}^f_k$ in hyperspace while preserving contextual information in given images by enabling it to be trainable in a gradient descent-based self-supervised way. The encoder model $\vec{\phi}$ consists of two major modules: $E_{ext}\colon\mathbb{R}^z\to\mathbb{R}^{z'}$ where $z > z'$ and $E_{gen}\colon\mathbb{R}^{z'}\to\mathbb{R}^D$. For a given visual feature vector $\vec{f}_k$, $\vec{\phi}$ is applied as $\vec{h}^f_k = \vec{\phi}(\vec{f}_k) = E_{gen}(E_{ext}(\vec{f}_k))\in\mathbb{R}^D$. $E_{ext}$ acts as important visual feature extractor and $E_{gen}$ acts as a mapper to hyperspace. To harness general context representation, we limit the dimensionality by $z'$, which prevents overfitting. We train the function $\vec{\phi}$ using the following loss function: $\mathcal{L}_{Enc} = \mathcal{L}_c + \lambda_{rec}\mathcal{L}_{rec}$ where $\lambda_{rec}\in\mathbb{R}$ indicates balance coefficient. $\mathcal{L}_c$ uses pairs of $M$ object images $I_k$ and corresponding pseudo-labels $\tilde{y}_k$ that are generated using the pre-trained object detection model $f_{obj}$ as shown in~\autoref{eq:losscross} where $C\in\mathbb{R}^{D\times c}$ stands as class hypervectors with $c$ pseudo-classes. A simple linear layer module $E_{rec}\colon \mathbb{R}^D\to\mathbb{R}^z$ is introduced in order to compute the reconstruction loss $\mathcal{L}_{rec}$ as shown in~\autoref{eq:lossrec} to force the model $\vec{\phi}$ to preserve original features' information in hyperspace.
\vspace{-1mm}
\begin{equation}\label{eq:losscross}
\mathcal{L}_{c} = \sum_{k}\textit{CrossEntropy}(\textit{softmax}(\vec{\phi}(\vec{f}_k))^TC), \tilde{y}_k)
\end{equation}
\vspace{-3mm}
\begin{equation}\label{eq:lossrec}
\mathcal{L}_{rec} = \frac{1}{M}\sum_{k}{\left \| \vec{f}_k - E_{rec}(\vec{\phi}(\vec{f}_k)) \right \|^2}
\end{equation}

\subsubsection{Hyperdimensional Spatial-aware Encoding}

Given global feature hypervector $\vec{h}^f_{glob}$ and local feature hypervectors $\vec{h}^f_{k}$ with their corresponding object positions $p_k=(x_k, y_k)$ final hyperdimensional representation $\vec{H}$ of the given image $I$ is computed using HDC operations. First, to encode position information $p_k$, positional base hypervectors $\vec{B}_X$ and $\vec{B}_Y$ are randomly sampled from a normal distribution $\{\mathcal{N}(0, 1)\}^{D}$. With the randomly sampled positional base hypervectors $\vec{B}_X$ and $\vec{B}_Y$, each $x_k$ and $y_k$ are projected to the hyperspace using $\vec{B}_X$ and $\vec{B}_Y$ respectively with $\exp^i(.)$ function where $i$ indicates imaginary unit ($i=\sqrt{-1}$). Each embedding of the x-axis and y-axis dimensional information is computed by $\vec{h}^X_k = \exp^i{\left (x_k \vec{B}_X\right )} = \begin{bmatrix} e^{i\vec{B}_{X_1} x_k} & e^{i\vec{B}_{X_2} x_k} & \cdots & e^{i\vec{B}_{X_D} x_k} \end{bmatrix} \in \mathbb{C}^D$ and $\vec{h}^Y_k = \exp^i{\left (y_k \vec{B}_Y\right )} = \begin{bmatrix} e^{i\vec{B}_{Y_1} y_k} & e^{i\vec{B}_{Y_2} y_k} & \cdots & e^{i\vec{B}_{Y_D} y_k} \end{bmatrix} \in \mathbb{C}^D$ respectively. Final positional hypervectors $\vec{h}^p_k = \vec{h}^X_k * \vec{h}^Y_k = e^{i\vec{B}_{X_j} x_k + i\vec{B}_{Y_j} y_k} \in\mathbb{C}^D$ are computed by combining the two hypervectors $\vec{h}^X_k$ and $\vec{h}^Y_k$ using the binding operation to associate both $(x, y)$-axis dimensional information.

Additionally, we can introduce a new hyperparameter length scale $w$. The length scale acts like a factor that controls the standard deviation of $\vec{B}_X$ and $\vec{B}_Y$ by being placed in the $\exp^i(.)$ function as $\exp^{i/w}(.)$. With the smaller $w$, it affects $\vec{B}_X$ and $\vec{B}_Y$ are sampled from a normal distribution with higher standard deviation $\{\mathcal{N}(0, \frac{1}{w})\}^{D}$ creating sparse representation of positional hypervectors. While larger $w$ affects $\vec{B}_X$ and $\vec{B}_Y$ are sampled from smaller standard deviations making representation of positional hypervectors more dense. Thus, by controlling $w$, we can adjust the magnitude of the association of spatial information.

Now, to have the final hyperdimensional representation, positional hypervectors are combined with the visual feature hypervectors that are retrieved from the global and local visual features extraction process. Each local visual feature vector $\vec{h}^f_k\in\mathbb{R}^D$ is paired with the corresponding positional hypervector $\vec{h}^p_k\in\mathbb{C}^D$. Each pair $(\vec{h}^f_k, \vec{h}^p_k)$ is associated with each other resulting in a single hypervector by the following binding operation: $\vec{h}^f_k * \vec{h}^p_k \in \mathbb{C}^D$ represents visual and positional information. Lastly, Spatial embedding by bundling $\vec{h}^f_k * \vec{h}^p_k$ for all $k=1, 2, \cdots, N$ and global embedding by bundling $\vec{h}^f_{glob}$ with $\sum_{k}{\vec{h}^f_k * \vec{h}^p_k}$ are conducted resulting the final hyperdimensional representation $\vec{H} = \vec{h}^f_{glob} + \sum_{k}{\vec{h}^f_k * \vec{h}^p_k}$.

Furthermore, we can utilize Symbolic Training shown in~\autoref{sec:symbolictraining} where we merge separate symbolic representations into a single hypervector and optimize by giving weights to each symbolic hypervector to have a user desire hyperdimensional representations: $\vec{H} = \eta_{glob}\vec{h}^f_{glob} + \sum_{k}{\eta_k\vec{h}^f_k * \vec{h}^p_k}$ where $\eta_{glob}\in\mathbb{R}$ indicates weight on global features and $\eta_{k}\in\mathbb{R}$ indicates weight on each local features. Relatively higher weight $\eta$ leads to focused hyperdimensional representation for those highly weighted symbols. Simply manipulating $\eta_k$, we can have a new customized representation that can be used for conditional image retrieval without any heavy and time-consuming gradient-based optimization. Possible ways to automate assigning $\eta_k$ during hashing massive amounts of images are: by the size of bounding box $BB_k$, confidence score from the object detection model $f_{obj}$, etc. In this paper, we focus on the evaluation of manipulating query images' representation to have conditional image retrieval thus, we set the same amount of $1=\eta_{glob} = \eta_k, \forall k$ during the hashing retrieval set.

\subsubsection{Multilinear Hyperplane Hashing Optimization}

We explored that by utilizing HDC operations in hyperspace, hyperdimensional representation $\vec{H}_n\in\mathbb{C}^D$ that well represents both spatial-aware local context and global context of a given image $I_n\in\mathcal{I}$ can be driven. However, due to the high dimensionality $D\gg z$ and the high precision for representing each element $\vec{H}_{ni}\in\mathbb{C}$ it is infeasible to adapt the hyperdimensional representations in fast image retrieval task directly. Thus, it is necessary to have a hash function $\vec{f}_h\colon \vec{H} \mapsto \vec{H}'$ which maps given hyperdimensional representation $\vec{H}$ to a compact $L$-bit hash representation $\vec{H}'\in \{-1, +1\}^L$ that well preserves relationships in hyperspace within low-dimensional hamming space.

To have a well-performing hash function $\vec{f}_h$, we utilize the locality-sensitive hashing (LSH) method by making it trainable in a gradient descent way. Among various different variations of LSH, we target to optimize random multilinear hyperplane hashing~\cite{liu2016multilinear} method that is specifically designed to preserve relationships in cosine similarity. The model initialization is similarly conducted as the previous work by randomly sampling $p_{ij} \sim \mathcal{N}(0, 1)$. Each $\vec{p}_i\in\mathbb{R}^{2D}$ represents a randomly sampled hyperplane that lies on the hyperspace dimensionality of $2D$. Notably, the hyperplanes lie on $2D$-dimensional space, not $D$, as a result of placing the hyperdimensional representation $\vec{H}_n$ which consists of complex numbers to real number space $\mathbb{R}^{2D}$ by concatenating real and imaginary parts $\Re(\vec{H}_n)^\frown\Im(\vec{H}_n) \in \mathbb{R}^{2D}$. Each hyperplane assigns a single bit value to each hyperdimensional data point by dividing them into two. Using a function $\textit{sign}(.)$ that returns $+1$ if a given value $x\in\mathbb{R}$ is larger or equal to $0$ otherwise returns $-1$, this can be represented as $\vec{H}'_{ni} = \textit{sign}(\vec{p}_i\cdot \Re(\vec{H}_n)^\frown\Im(\vec{H}_n))\in\{+1, -1\}$. As two hypervectors are divided into more common sides by hyperplanes they are considered to be also similar in the original hyperspace in terms of cosine similarity.

To optimize randomly sampled hyperplanes from a normal distribution, we generalized our hashing function as $f_h(H) = \tanh{\left ( HP^T + b \right )}\in [-1, +1]^{M\times L}$ where $H\in\mathbb{R}^{M\times 2D}$ indicates given $M$ concatenated hyperdimensional representations $H_n = \Re(\vec{H}_n)^\frown\Im(\vec{H}_n)$, $P\in\mathbb{R}^{L\times 2D}$ indicates $L$ hyperplanes, and $b\in\mathbb{R}^L$ indicates bias. We used $\tanh(.)$ function instead of $\textit{sign}(.)$ to avoid indifferentiable characteristic of $\textit{sign}(.)$. Thus, the final $L$-bit binary representation needs to be retrieved by $B = \textit{sign}(H')\in\{+1, -1\}^{M\times L}$ where $H' = f_h(H)$. Finally, we introduce the loss function that is formulated as the following:
\vspace{-1mm}
\begin{equation}\label{eq:lossfunc}
\mathcal{L}_{Hyper} = \lambda_{mse}\mathcal{L}_{mse} + \lambda_{w}\mathcal{L}_{w} + \lambda_{q}\mathcal{L}_{q} + \lambda_{u}\mathcal{L}_{u} + \lambda_{o}\mathcal{L}_{o}
\end{equation}\vspace{-4mm}

The loss function that is shown in \autoref{eq:lossfunc} consists of 5 loss terms: mean square error (MSE) loss $\mathcal{L}_{mse}$, w-shape loss $\mathcal{L}_{w}$, quantization loss $\mathcal{L}_{q}$, uniform loss $\mathcal{L}_{u}$, and order loss $\mathcal{L}_{o}$. Each loss term has its balance coefficients: $\lambda_{mse}$, $\lambda_{w}$, $\lambda_{q}$, $\lambda_{u}$, and $\lambda_{o}$. These terms are formulated in order to resolve four issues:

\begin{table*}[!htb]
\begin{minipage}[t]{.8\textwidth}
\centering
        \begin{adjustbox}{width=0.97\textwidth}
        \begin{tabular}{c|c|c|c|c|c|c|c}
        \toprule
        \multirow{2}{*}{\textbf{Methods}} & \multirow{2}{*}{\textbf{References}} & \multicolumn{3}{c|}{\textbf{CIFAR10}} & \multicolumn{3}{c}{\textbf{MS COCO}} \\
        \cline{3-8}
        & & \rule{0pt}{2.6ex}16 bits & 32 bits & 64 bits & 16 bits & 32 bits & 64 bits \\ 
        \midrule
        \midrule
        AGH~\cite{liu2011hashing} & ICML11 & 0.333 & 0.357 & 0.358 & 0.596 & 0.625& 0.631 \\
        ITQ~\cite{gong2012iterative} & TPAMI12 & 0.305 & 0.325 & 0.349 & 0.598 & 0.624 & 0.648 \\
        DGH~\cite{liu2014discrete} & NeurIPS14 & 0.335 & 0.353 & 0.361 & 0.613 & 0.631 & 0.638 \\
        SGH~\cite{dai2017stochastic} & ICML17 & 0.435 & 0.437 & 0.433 & 0.594 & 0.610 & 0.618 \\
        BGAN~\cite{song2018binary} & AAAI18 & 0.525 & 0.531 & 0.562 & 0.645 & 0.682 & 0.707 \\
        GreedyHash~\cite{su2018greedy} & NeurIPS18 & 0.448 & 0.473 & 0.501 & 0.582 & 0.668 & 0.710 \\
        DVB~\cite{shen2019unsupervised} & IJCV19 & 0.403 & 0.422 & 0.446 & 0.570 & 0.629 & 0.623 \\
        TBH~\cite{shen2020auto} & CVPR20 & 0.497 & 0.524 & 0.529 & 0.706 & 0.735 & 0.722 \\
        CIB~\cite{qiu2021unsupervised} & IJCAI21 & 0.547 & 0.583 & 0.602 & 0.737 & 0.760 & 0.775\\
        HAMAN~\cite{ma2022improved} & IJCAI22 & - & - & - & 0.722 & 0.775 & 0.787 \\
        NSH~\cite{yu2022learning} & IJCAI22 & 0.706 & 0.733& 0.756 & 0.746 & 0.774 & 0.783 \\
        FSCH~\cite{cao2023unsupervised} & TCSVT23 & \textbf{0.876} & 0.912 & 0.926 & 0.760 & 0.787 & 0.799 \\
        \midrule
        na\"ive (DINOv2 + LSH) &  & 0.316 & 0.450 & 0.599 & 0.479 & 0.557 & 0.658 \\
        \midrule
        \textbf{\NeuroHash ($w=1.0$)} & \textbf{Ours} & 0.839 & \textbf{0.937} & 0.927 & \textbf{0.785} & \textbf{0.878} & \textbf{0.904} \\
        \textbf{\NeuroHash ($w=10.$)} & \textbf{Ours} & 0.827 & 0.912 & \textbf{0.945} & 0.780 & 0.873 & 0.903 \\
        \bottomrule
    \end{tabular}
    \end{adjustbox}
    \vspace{1mm}
    \caption{mAP@5K results for different methods on datasets CIFAR10 and MS COCO.}
    \label{tab:map}
    \vspace{-6.5mm}

\end{minipage}%
\begin{minipage}[t]{0.20\textwidth}
\centering
    \vspace{-37mm}
    \begin{adjustbox}{width=0.95\textwidth}
    \begin{tabular}{c|c}
        \toprule
        \textbf{Without}  & \textbf{mAP@5K} \\
        \midrule
        \midrule
        $\mathcal{L}_{Hyper}$ & 0.623 \\
        $\mathcal{L}_{mse}$ & 0.505 \\
        $\mathcal{L}_{w}$ & 0.940 \\
        $\mathcal{L}_{q}$ & 0.933 \\
        $\mathcal{L}_{u}$ & 0.894 \\
        $\mathcal{L}_{o}$ & 0.942 \\
        \midrule
        \textbf{Full Model} & \textbf{0.945} \\
        \bottomrule
    \end{tabular}
    \end{adjustbox}
    \vspace{1mm}
    \caption{Ablation studies on Multilinear Hyperplane Hashing Optimization of \NeuroHash using the mAP@5K metric on the CIFAR10 dataset for 64 bits with the scale factor of $w = 10$. The test is done by removing each loss term including the case where using only random hyperplanes ($\mathcal{L}_{Hyper}$).}
    \label{tab:ablation}
    \vspace{-6.5mm}
\end{minipage}
\end{table*}

\paragraph{Loss term for numerical correspondence.} First of all, the MSE loss term $\mathcal{L}_{mse}$ is used to match the numerical similarity between cosine similarity in hyperspace and $L$-precision hamming distance as shown in \autoref{eq:mseloss}.
\vspace{-1mm}
\begin{equation}\label{eq:mseloss}
\mathcal{L}_{mse} = \frac{1}{M^2}\sum_{1\leq i, j \leq M}{\left \| \frac{H_{i}H_{j}^T}{\|H_{i}\|\|H_{j}\|} - \frac{H'_{i}{H'}_{j}^T}{L} \right \|^2}
\end{equation}\vspace{-1mm}

To match the hamming distance value with the similarity value, we used reversed hamming distance: $L - \sum_{1\leq k\leq L}{|B_{i_k} - B_{j_k}|} = B_{i}B_{j}^T\approx H'_{i}{H'}_{j}^T$. We further adjusted the range by $-1\leq \frac{H'_{i}{H'}_{j}^T}{L} \leq +1$ as the same as the range of cosine similarity value.

\paragraph{Loss terms for limited representation.} Due to the low precision bits representation, distance is also extremely discrete which makes indistinguishable distances between many images. To tackle this issue, we set an assumption that in most cases, boundary distance is not placed among distances that are located on either side of the edges -- either distance is very close or very far. Based on this assumption, we applied another loss term we named w-shape loss presented in \autoref{eq:wloss}. This loss function gives more penalty for the distances that are more closely located in the center. In the same context of low precision and low dimensionality, it also can cause limited unique representations. To avoid such representation collapsing, we also introduced uniform loss as shown in \autoref{eq:uloss}. Note that $\textbf{1}_L$ indicates $L$-dimension vector consists of ones. $(H'_i\textbf{1}_L^T)^2$ will be closer to zero as the number of $+1$s and the number of $-1$s gets closer which forces the hashing model to generate a uniform number of binary representations.
\vspace{-3mm}
\begin{equation}\label{eq:wloss}
\mathcal{L}_{w} = \frac{1}{M^2}\sum_{1\leq i, j \leq M}{\left ( \frac{H'_{i}{H'}_{j}^T}{L} + 1 \right )^2 \left ( \frac{H'_{i}{H'}_{j}^T}{L} - 1 \right )^2}
\end{equation}
\begin{equation}\label{eq:uloss}
\mathcal{L}_{u} = \frac{1}{M}\sum_{1\leq i \leq M}{(H'_i\textbf{1}_L^T)^2}
\end{equation}

\paragraph{Loss term for learning binary representations.} Next, since we are using $\tanh(.)$ function instead of $\textit{sign}(.)$ we also applied quantization loss as shown in \autoref{eq:qloss}. This loss term helps the hashing model $f_h$ to generate representations that are close to the binary representation such as $H'_{ij}\approx \textit{sign}(H'_{ij})$.
\begin{align}\label{eq:qloss}
\mathcal{L}_{q} = \frac{1}{NL}\sum_{1\leq i \leq M}{\sum_{1 \leq j \leq L}{\left ( H'_{ij} - \textit{sign}(H'_{ij}) \right )^2}} \nonumber \\ = \frac{1}{NL}\sum_{1\leq i \leq M}{\sum_{1 \leq j \leq L}{\left ( H'_{ij} - B_{ij} \right )^2}} 
\end{align}

\paragraph{Loss term for reversed relative order.} For the last loss term, we consider the relative orders between hyperdimensional representation pairs. It targets to preserve the order of ranking that each $H_i$ has to all the other $H_j$. In other words, for the pair $(i, j)$, the number of $k\in \{1, 2, \cdots, M\}$ that satisfies $\delta(H_i, H_j) > \delta(H_i, H_k)$, should be similar as the number of $k\in \{1, 2, \cdots, M\}$ that satisfies $\delta(H'_i, H'_j) > \delta(H'_i, H'_k)$. \autoref{eq:oloss} shows a loss term that gives a penalty in such cases that the number of $k$ that satisfies the above condition is not satisfied we named order loss. The function $C(.,.)$ determines order reversed cases. If the order is reduced, it gives a larger penalty to the larger hamming distance and if the order is increased, it gives a larger penalty to the smaller hamming distance otherwise, it gives zero penalty.

\vspace{-2mm}
\begin{equation}\label{eq:oloss}
\mathcal{L}_{o} = \frac{1}{M^2}\sum_{1\leq i, j \leq M}{C(i, j) \begin{bmatrix} \left ( 1 - \frac{{H'}_i {H'}_j^T}{L} \right )^2 & \left ( 1 + \frac{{H'}_i {H'}_j^T}{L} \right )^2 \end{bmatrix}^T }
\end{equation}
\vspace{-2mm}
\begin{tiny}
\begin{equation}
C(i, j) = \begin{cases}
    \begin{bmatrix} 1 & 0 \end{bmatrix} & \text{if } j \in \left \{ v \middle| |\{k | \frac{H'_i {H'}_v^T}{L} > \frac{H'_i {H'}_k^T}{L} \}| < |\{k | \frac{H_i H_v^T}{L} > \frac{H_iH_k^T}{L} \} | \right \} \\
    \begin{bmatrix} 0 & 1 \end{bmatrix} & \text{if } j \in \left \{ v \middle| |\{k | \frac{H'_i{H'}_v^T}{L} > \frac{H'_i{H'}_k^T}{L} \}| > |\{k | \frac{H_iH_v^T}{L} > \frac{H_iH_k^T}{L} \} | \right \} \\
    \begin{bmatrix} 0 & 0 \end{bmatrix} & \text{otherwise}
  \end{cases} \nonumber
\end{equation}
\end{tiny}

\section{Experiments}

\subsubsection{Experiment settings}

\begin{table*}[ht]
    \centering
    \vspace{-3mm}
    \begin{adjustbox}{width=1.\textwidth,center=\textwidth}
    \begin{tabular}{c|c|c|c|c|c|c|c|c|c|c|c|c}
        \toprule
        \multirow{2}{*}{\textbf{Scale Factor ($\boldsymbol{w}$)}}  & \multicolumn{3}{c|}{$\boldsymbol{mAP@5K_{r=0.1}}$} & \multicolumn{3}{c|}{$\boldsymbol{mAP@5K_{r=0.2}}$} & \multicolumn{3}{c|}{$\boldsymbol{mAP@5K_{r=0.3}}$} & \multicolumn{3}{c}{$\boldsymbol{mAP@5K_{r=0.4}}$} \\
        \cline{2-13}
        & \rule{0pt}{2.6ex} 16 bits & 32 bits & 64 bits & 16 bits & 32 bits & 64 bits & 16 bits & 32 bits & 64 bits & 16 bits & 32 bits & 64 bits \\
        \midrule
        \midrule
        $w=0.1$ & \textbf{0.698} & \textbf{0.757} & \textbf{0.776} & \textbf{0.926} & \textbf{0.945} & \textbf{0.952} & \textbf{0.976} & \textbf{0.981} & \textbf{0.983} & \textbf{0.991} & \textbf{0.992} & \textbf{0.993} \\
        $w=1.0$ & 0.626 & 0.632 & 0.634 & 0.885 & 0.888 & 0.889 & 0.964 & 0.965 & 0.965 & 0.988 & 0.988 & 0.988 \\
        $w=10.$ & 0.622 & 0.631 & 0.632 & 0.882 & 0.887 & 0.887 & 0.962 & 0.964 & 0.964 & 0.988 & 0.988 & 0.988 \\
        \bottomrule
    \end{tabular}
    \end{adjustbox}
    \vspace{1mm}
    \caption{Evaluation results on MS COCO dataset with newly proposed metric $mAP@K_r$ specifically designed to evaluate spatial-ware image retrieval performance.}
    \label{tab:mapr}
    \vspace{-6mm}
\end{table*}

\begin{figure*}[ht]
    \centering
    \includegraphics[width=1.\textwidth]{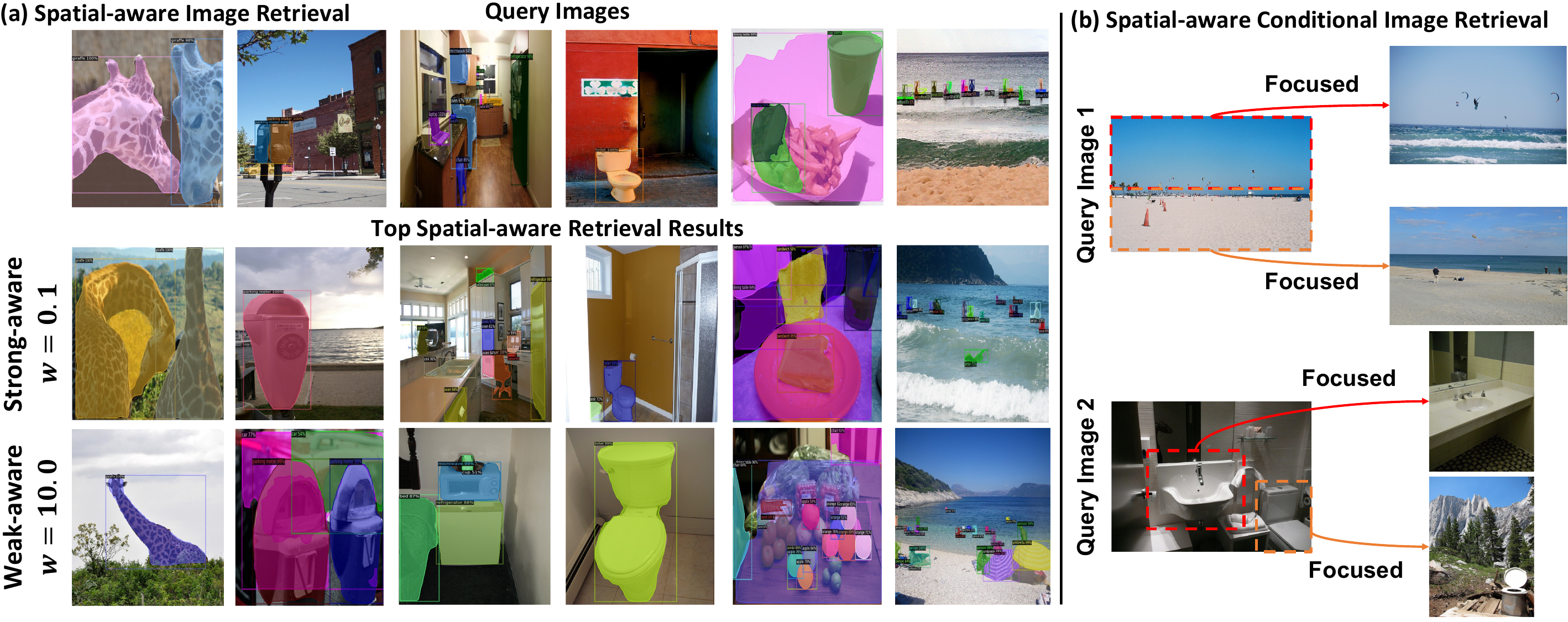}
    \vspace{-6mm}
    \caption{\textbf{Qualitative evaluation of \NeuroHash on  (a) Spatial-aware Retrieval and (b) Spatial-aware Conditional Image Retrieval}.}
    \vspace{-4mm}
    \label{fig:ijicai2024_exp}
\end{figure*}

\begin{enumerate}
    \item \textbf{Implementation Details} For the object detection model $f_{obj}(.)$ we used Detectron 2~\cite{wu2019detectron2} and for the image embedding model $\vec{\phi}_{vis}(.)$ we used DINOv2 ViT-g/14 model~\cite{oquab2023dinov2}. Since the ViT-g/14 model uses a patch size of 14, it is necessary to transform the image size into a multiplier of 14 for both width and height. It is implemented as transforming a given image $I_k$ of size $(w, h)$ to $((\lfloor{w / 14}\rfloor + 1)\times 14, (\lfloor{h / 14}\rfloor + 1)\times 14)$. For the hypervectors, we used the dimensionality of $D=10,000$, which is commonly selected in the HDC domain. 
    
    \item \textbf{Evaluation Metrics} In order to thoroughly evaluate our proposed method and compare it to conventional baselines, we used mAP (mean Average Precision), a widely accepted metric for evaluating retrieval performance. This metric calculates the average precision (AP) for a given query and a ranked list of returned results, where mAP is determined by averaging the AP values across all queries. In our evaluation, we follow the latest convention and use mAP@5000 for CIFAR-10 and MS COCO. Higher mAP values indicate better overall performance. 

    In addition to conventional evaluation metrics, we introduce a novel metric called \textit{mAP@K${r}$} to measure the effectiveness of our proposed spatial-aware conditional image retrieval. This metric represents a spatial-aware version of mAP and evaluates whether the coordinates of objects in the query image align with those in the retrieved image. This alignment is determined by calculating the Euclidean distance between the ground truth object coordinates and those of the retrieved image. The parameter $r$ defines the metric's spatial sensitivity by determining correct retrieval for two objects' $i$ and $j$ having the same class using $(\frac{x_i}{w_i} - \frac{x_j}{w_j})^2 + (\frac{y_i}{h_i} - \frac{y_j}{h_j})^2 \leq r^2$ where $x_i, y_i$ and $x_j, x_j$ represent each object's coordinates in their image and $w_i, h_i$ and $w_j, h_j$ indicate each image's dimensionality. Consequently, a higher value of $r$ results in a more lenient evaluation of whether the retrieved object contains similar objects at the same location. As the coordinate information is crucial for our proposed framework, we performed evaluations exclusively on the MS COCO dataset, using mAP@K${r=0.1}$, mAP@K${r=0.2}$, mAP@K${r=0.3}$ and mAP@K${r=0.4}$.
\end{enumerate}

\subsubsection{Datasets}

\begin{enumerate}
    \item \textbf{\textit{MS-COCO}} ~\cite{lin2014microsoft} has 82,783 training samples and 40,504 validation samples, with each image annotated with one or more labels from a pool of 91 categories. In this study, we follow the previous research~\cite{cao2023unsupervised}, a subset of 122,218 images from 80 categories is used. Within this subset, a random sample of 5,000 images is referred to as the query dataset, while the remaining images form the retrieval set. In particular, MS COCO stands out from other datasets due to the inclusion of ground truth bounding box information, providing a unique opportunity to assess the extent to which our proposed method captures local information using our proposed metric \textit{mAP@K${r}$}.
    
    \item \textbf{\textit{CIFAR-10}} ~\cite{krizhevsky2009learning} involves 60,000 images distributed across 10 categories, with each class containing 6,000 images. Following the earlier study ~\cite{cao2023unsupervised}, we randomly chose 100 images from each class to form the query dataset, amounting to a total of 1000 images. Subsequently, we utilized the remaining images for retrieval purposes.
\end{enumerate}

\subsubsection{Evaluation on Weak-spatial-aware Image Retrieval}

First, we evaluated our \NeuroHash on \textit{Weak-spatial-aware} image retrieval case with other hashing methods including current state-of-the-art models. On \textit{Weak-spatial-aware} image retrieval, we focus on conventional image retrieval metric mAP@K which evaluates without spatial alignment of each object shown in the images. In this test, we gave high-scale factors $w=1., 10.$ to make it less focused on spatial information. \autoref{tab:map} shows mAP@5K results for different methods including na\"ive approach, which uses our backbone model DINOv2 ViT-g/14 directly with conventional hashing algorithm LSH, and ours with different scale factors $w$ on two different datasets: CIFAR10 and MS COCO. As reported in the table, our \NeuroHash shows strong results on the non-spatial aware image retrieval metric by outperforming other methods in most of the cases by up to 13.14\% improvement.

\subsubsection{Evaluation on Strong-spatial-aware Image Retrieval}

To ensure the efficacy of our proposed \NeuroHash on spatial-aware conditional image retrieval task, we conducted image retrieval evaluation on \textit{Strong-spatial-aware} image retrieval case which aims to retrieve images with similar object positioning. In this evaluation, we use our proposed mAP@5K$_r$ metric with $r=0.1, 1.0, 10.0$. Since the mAP@5K$_r$ metric requires ground truth labels on positional information, we used only the MS COCO dataset which provides positional annotations that other datasets are not providing. As presented in~\autoref{tab:mapr}, we can observe that by decreasing the scale factor $w$, we achieve a higher mAP@5K$_r$ score which indicates higher spatial awareness during the image retrieval.

\subsubsection{Evaluation on Conditional Image Retrieval}

In this conditional image retrieval section, we visually demonstrate spatial-aware image retrieval and conditional retrieval of our \NeuroHash shown in~\autoref{fig:ijicai2024_exp}. On \autoref{fig:ijicai2024_exp}.(a) shows the effect of controlling $w$ to control between weak and strong spatial awareness. When $w=0.1$, the retrieved images present higher positional alignments to the objects in the query image compared to $w=10.0$. On \autoref{fig:ijicai2024_exp}.(b), we showcase two conditional image retrieval when retrieval image set is hashed with $w=0.1$ and $\eta_i=1$. For the query image 1, we set $\eta_i=10$ for the objects located in the focusing region and $\eta_i=1$ for outside of the region. For the query image 2, we set $\eta_i=10$ for a specific object and $\eta_i=1$ for the others. We can observe the retrieved images are highly focused in terms of positional and visual matching on those regions or objects that have higher $\eta_i$.

\subsubsection{Ablation Study on Multilinear Hyperplane Hashing Opt.}

\autoref{tab:ablation} shows an ablation study on our model. The results indicate that all loss metrics are necessary for effective hash value generation on the CIFAR10 dataset using the mAP@5K metric. As shown, the full model achieved the highest score.

\section{Conclusions}

In this paper, we propose \NeuroHash a completely novel approach to hashing images in a neuro-symbolic way that enables spatial-aware hashing and conditional image retrieval. Experiments on well-known datasets for image retrieval performance benchmarking validate the efficacy of our work. In future work, we aim to evolve a more versatile approach capable of embedding various types of information, including temporal information. This future work seeks to broaden the scope of our neuro-symbolic framework, fostering its application in diverse domains beyond spatial-aware image retrieval.

\section*{Acknowledgements}
This work was supported in part by the DARPA Young Faculty Award, the National Science Foundation (NSF) under Grants \#2127780, \#2319198, \#2321840, \#2312517, and \#2235472, the Semiconductor Research Corporation (SRC), the Office of Naval Research through the Young Investigator Program Award, and Grants \#N00014-21-1-2225 and \#N00014-22-1-2067. Additionally, support was provided by the Air Force Office of Scientific Research under Award \#FA9550-22-1-0253, along with generous gifts from Xilinx and Cisco.

\bibliographystyle{IEEEtranS}
\bibliography{refs}

\end{document}